\documentclass[sigconf]{acmart}

\newcommand{\eg}{\textit{e.g.}}
\usepackage[capitalize]{cleveref}
\crefname{section}{Sec.}{Secs.}
\Crefname{section}{Section}{Sections}
\Crefname{table}{Table}{Tables}
\crefname{table}{Tab.}{Tabs.}

\usepackage[page]{appendix}
\usepackage{balance}
\usepackage{array}
\usepackage{multirow}
\newcolumntype{L}[1]{>{\raggedright\arraybackslash}p{#1}}
\newcolumntype{C}[1]{>{\centering\arraybackslash}p{#1}}

\copyrightyear{2025}
\acmYear{2025}
\setcopyright{acmlicensed}
\acmConference[MM '25] {Proceedings of the 33rd ACM International Conference on Multimedia}{October 27--31, 2025}{Dublin, Ireland.}
\acmBooktitle{Proceedings of the 33rd ACM International Conference on Multimedia (MM '25), October 27--31, 2025, Dublin, Ireland}
\acmISBN{979-8-4007-2035-2/2025/10}
\acmDOI{10.1145/3746027.3755369}

\acmSubmissionID{3465}
\begin{document}

\title{Color Me Correctly: Bridging Perceptual Color Spaces and Text Embeddings for Improved Diffusion Generation}

\author{Sung-Lin Tsai}
\orcid{0009-0002-1859-2406}
\affiliation{
    \institution{National Yang Ming Chiao Tung University}
    \city{Hsinchu}
    \country{Taiwan}
}
\email{tsai412504004.ee12@nycu.edu.tw}

\author{Bo-Lun Huang}
\orcid{0009-0003-9804-5046}
\affiliation{
  \institution{National Yang Ming Chiao Tung University}  
  \city{Hsinchu}
  \country{Taiwan}
}
\email{kevin503.ee12@nycu.edu.tw}

\author{Yu-Ting Shen}
\authornote{Contribute equally to this work.}
\orcid{0009-0008-0146-8154}
\affiliation{
  \institution{National Yang Ming Chiao Tung University}
  \city{Hsinchu}
  \country{Taiwan}
}
\email{yuting89830.cs11@nycu.edu.tw}

\author{Cheng-Yu Yeo}
\authornotemark[1]
\orcid{0009-0003-8474-6140}
\affiliation{
  \institution{National Yang Ming Chiao Tung University}
  \city{Hsinchu}
  \country{Taiwan}
}
\email{boyyeo123.ee12@nycu.edu.tw}

\author{Chiang Tseng}
\authornotemark[1]
\orcid{0009-0009-7750-1456}
\affiliation{
  \institution{National Yang Ming Chiao Tung University}
  \city{Hsinchu}
  \country{Taiwan}
}
\email{chiang.ee11@nycu.edu.tw}

\author{Bo-Kai Ruan}
\authornotemark[1]
\orcid{0000-0002-9847-3628}
\affiliation{
  \institution{National Yang Ming Chiao Tung University}
  \city{Hsinchu}
  \country{Taiwan}
}
\email{bkruan.ee11@nycu.edu.tw}

\author{Wen-Sheng Lien}
\orcid{0009-0004-1107-5590}
\affiliation{
  \institution{National Yang Ming Chiao Tung University}
  \city{Hsinchu}
  \country{Taiwan}
}
\email{vincentlien.ii13@nycu.edu.tw}

\author{Hong-Han Shuai}
\authornote{Corresponding author.}
\orcid{0000-0003-2216-077X}
\affiliation{
  \institution{National Yang Ming Chiao Tung University}
  \city{Hsinchu}
  \country{Taiwan}
  }
\email{hhshuai@nycu.edu.tw}

\renewcommand{\shortauthors}{Sung-Lin Tsai et al.}

\begin{abstract}
Accurate color alignment in text-to-image (T2I) generation is critical for applications such as fashion, product visualization, and interior design, yet current diffusion models struggle with nuanced and compound color terms (e.g., \textit{Tiffany blue}, \textit{baby pink}), often producing images that are misaligned with human intent. Existing approaches rely on cross-attention manipulation, reference images, or fine-tuning but fail to systematically resolve ambiguous color descriptions. To precisely render colors under prompt ambiguity, we propose a \textbf{training-free} framework that enhances color fidelity by leveraging a large language model (LLM) to disambiguate color-related prompts and guiding color blending operations directly in the text embedding space. Our method first employs a large language model (LLM) to resolve ambiguous color terms in the text prompt, and then refines the text embeddings based on the spatial relationships of the resulting color terms in the CIELab color space. Unlike prior methods, our approach improves color accuracy without requiring additional training or external reference images. Experimental results demonstrate that our framework improves color alignment without compromising image quality, bridging the gap between text semantics and visual generation. All supplementary materials are available at \url{https://Sung-Lin.github.io/TintBench/}.
\end{abstract}



\begin{teaserfigure}
    \vspace{-15pt}
    \centering
    \includegraphics[width=1.0\linewidth]{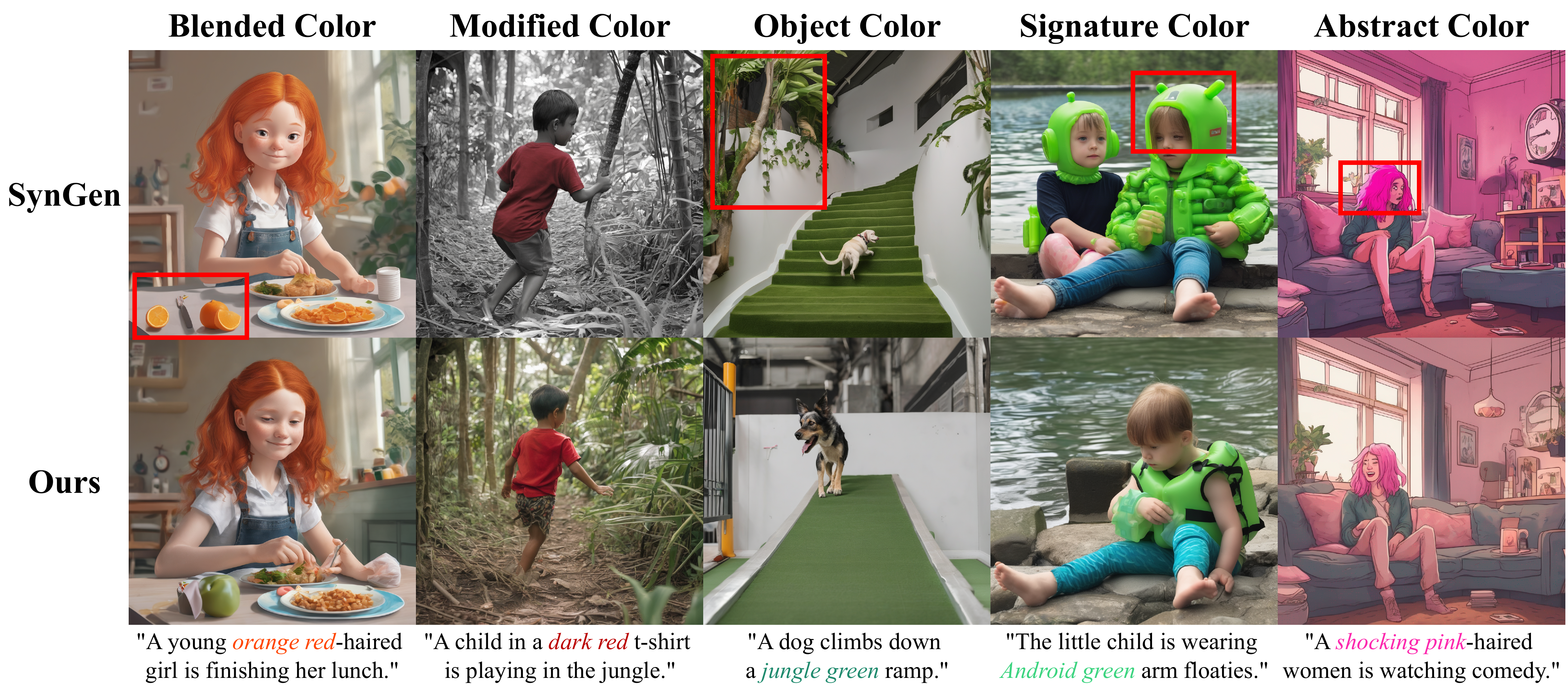}
    \caption{Examples of prompt-induced ambiguity. Top: baseline T2I diffusion (SynGen) outputs misinterpret color terms. Bottom: our disambiguation-guided method resolves these issues, yielding more accurate, semantically aligned results.}
    \Description{Examples of prompt-induced ambiguity.}
    \label{fig:teaser}
\end{teaserfigure}

\maketitle

\section{Introduction}
\label{sec:intro}
Color plays a crucial role in visual perception and aesthetics, significantly impacting domains such as fashion~\cite{han2023fashionsap}, product design~\cite{hou2025gencolor}, in-house decoration~\cite{shamoi2023towards}, and digital art~\cite{muratbekova2024color}. Accurate color alignment is essential for applications where color fidelity determines the outcome, such as e-commerce visualization, where customers expect generated product previews to match real-world colors precisely. Similarly, in virtual interior design, users may specify complex color schemes for furniture or walls, and any misinterpretation can lead to unrealistic renderings. However, current text-to-image (T2I) diffusion models often struggle with accurately rendering hues, particularly when interpreting compound color names such as \textit{lime green}, which can lead to imprecise color reproduction (e.g., generating \textit{plain green}) or object-color lexical confusion (e.g., generating limes).


To better address this issue, several approaches have been proposed to address color alignment in T2I generation. Early works such as Attend-and-Excite~\cite{chefer2023attend} and Divide \& Bind~\cite{li2023divide} explore cross-attention manipulation to enforce more faithful text-image correspondence, including color attributes. More recent methods introduce training-free solutions for color control. For example, ColorEdit~\cite{yin2024coloredit} applies reference-based color adjustments using cross-attention feature alignment, while Color-Style Disentanglement~\cite{agarwal2024training} leverages CIELab space feature separation to isolate color properties from luminance and style. Fine-tuning-based methods such as ColorPeel~\cite{butt2024colorpeel} train diffusion models to internalize new color embeddings, enabling precise color reproduction for specific shades. Meanwhile, LLM-based guidance frameworks like LLM-grounded Diffusion~\cite{lian2024llmgrounded} improve compositional reasoning in T2I synthesis but do not explicitly address color fidelity. While these approaches make significant strides, they often rely on reference images~\cite{yin2024coloredit, agarwal2024training}, training-intensive pipelines~\cite{butt2024colorpeel}, or indirect color manipulations~\cite{chefer2023attend, li2023divide, lian2024llmgrounded} that fail to resolve ambiguous or under-specified color descriptions systematically.

Specifically, a major challenge in addressing color ambiguity lies in interpreting complex color expressions and compound descriptors within natural language prompts. Many such expressions—\eg, \textit{cerulean blue}, \textit{dusty rose}, or \textit{warm taupe}—lack precise semantic grounding and may be interpreted differently depending on context or model internalization. While large diffusion models are trained on vast web data, they often struggle to render accurate colors when prompts include fine-grained or less conventional color names. This difficulty stems not from limited data, but from the semantic variability and ambiguity inherent in human color language, which is rarely standardized. As a result, text-to-image models often produce outputs that diverge from user intent when color terms are complex or linguistically nuanced.

As illustrated in \cref{fig:teaser}, text-to-image diffusion models frequently produce incorrect generations when the input prompt contains ambiguous or compound color terms. These failures typically arise from the model’s misinterpretation of modifiers or the holistic blending of complex expressions. For example, in the second column, the modifier \textit{dark} leads the model to apply a globally darker tone, diverging from the intended object-specific coloration. In the third column, the term \textit{jungle green} is misunderstood, prompting the model to insert jungle-like visual elements into the background instead of applying the correct hue. This motivates the need for more explicit disambiguation strategies that can bridge the gap between human color semantics and learned visual representations.

In this paper, we propose a novel approach to enhance color fidelity in T2I diffusion models via \textbf{Semantic Color Disambiguation with LLMs} and \textbf{Retrieval-Based Embedding Refinement for Color Representation}. Our method consists of two stages: First, an LLM refines ambiguous color descriptions by translating them into explicit, unambiguous terms for clearer intent interpretation. Second, we introduce a retrieval-based embedding refinement that interpolates between nearby basic color embeddings to yield a more precise target color representation. By mapping these interpolated terms to the CIELab space, we use surrogate color templates for numerical interpolation. This enables smooth, controllable color blending and embeddings that closely match the intended shade. Our approach integrates semantic understanding with visual accuracy, ensuring high-fidelity color rendering while preserving textual input. Unlike prior methods requiring reference images or tuning cross-attention, ours directly improves color awareness in a \textbf{training-free} manner—yielding significantly better color consistency and realism. Our main contributions are summarized as follows.
\begin{itemize}
    \item We identify critical challenges in achieving accurate color alignment in T2I synthesis, particularly with novel color terms. A new benchmark is established to systematically assess color fidelity across various T2I models, setting a standard for future evaluations.
    \item The method introduces two key modules: Semantic Color Disambiguation with LLMs to disambiguate color terms, enhancing semantic clarity; and Retrieval-Based Embedding Refinement for Color Representation, which applies semantic arithmetic of word embeddings based on precise surrogate color matching in the CIELab space. This training-free approach ensures efficient and scalable improvements in color fidelity for T2I synthesis.
    \item Extensive experiments demonstrate the superior performance of the proposed method. The results showcase significant improvements in color accuracy and consistency, validating the effectiveness of the approach in practical T2I generation tasks.
\end{itemize}

\section{Related Work}
\label{sec:related_work}

\subsection{Alignment with Attention Control}
Several techniques have manipulated cross-attention to improve text-to-image diffusion in various contexts. For instance, Prompt-to-Prompt~\cite{hertz2023prompttoprompt} preserves structure by partially injecting cross-attention maps, enabling fine-grained image edits. Attend-and-Excite~\cite{chefer2023attend} incorporate token-specific attention to avoid ``catastrophic neglect,'' thereby ensuring each prompt element is faithfully rendered. Likewise, Training-Free Layout Control~\cite{chen2024training} and MasaCtrl~\cite{cao2023masactrl} modify attention to respect layouts or maintain consistent appearances across edits. SynGen~\cite{rassin2023linguistic} aligns attention maps with grammatical structure to mitigate color-attribute swaps. On the other hand, a recent line of studies focuses on controlling color without retraining: Color-Style Disentanglement~\cite{agarwal2024training} transfers reference image color distributions in the CIELab space, disentangling color from luminance. ColorEdit~\cite{yin2024coloredit} aligns cross-attention features with a given color sample, enforcing color fidelity. While enhancing color rendering, these methods often rely on reference images or extensive attention tuning. In contrast, our approach better aligns the compound color without reference images or manual attention alignment.

\begin{figure}
    \centering
    \includegraphics[width=\linewidth]{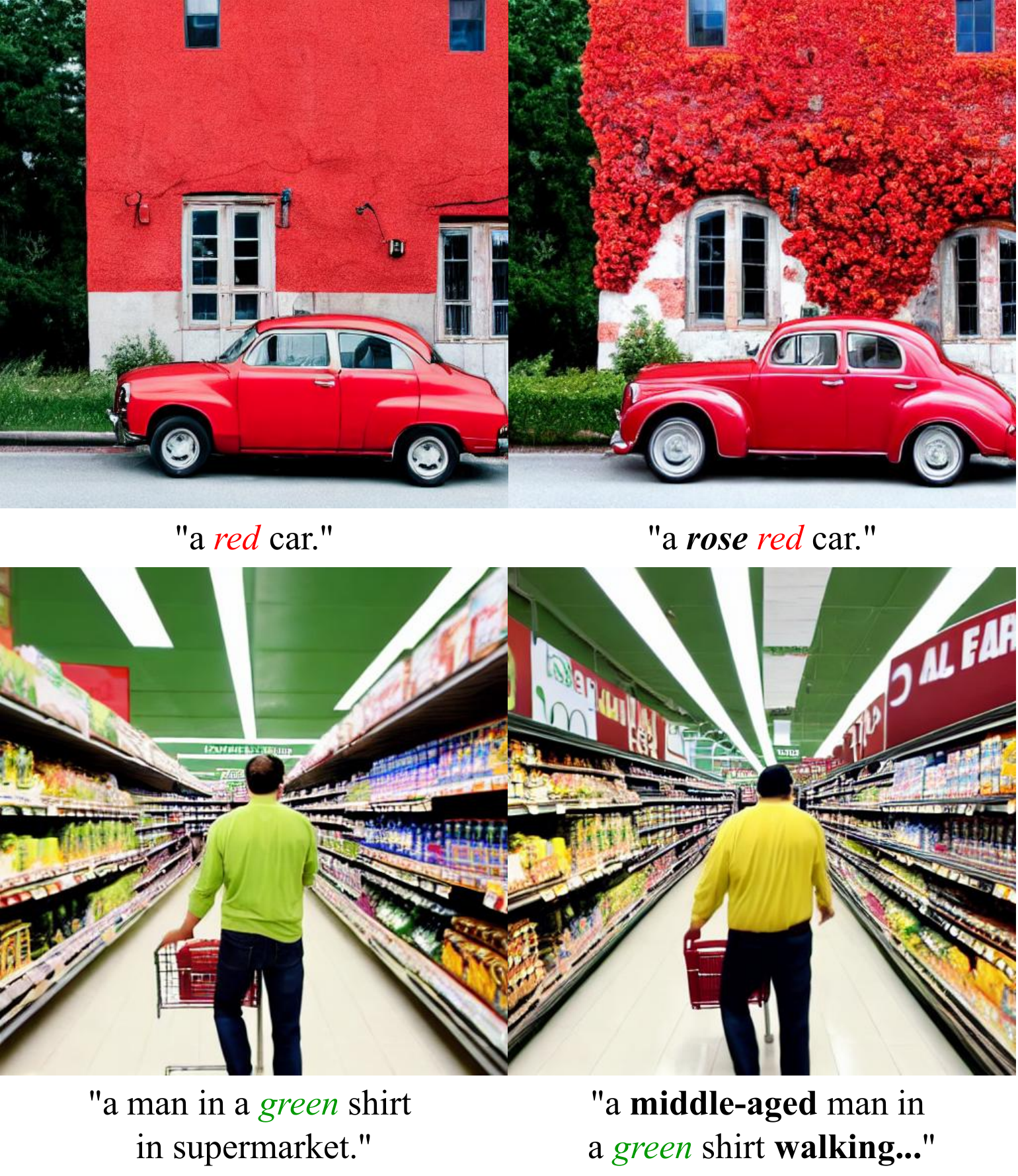}
    \caption{Failure cases of T2I diffusion models when processing prompts containing ambiguous color terms. Top: the term \textit{rose red} is misinterpreted, causing the model to generate multiple rose flowers instead of the intended color. Bottom: when additional descriptive details are added to the prompt (\eg \textit{middle-aged, walking}), the model incorrectly renders a green shirt as yellow.}
    \Description{Failure cases of T2I diffusion models when processing prompts containing ambiguous color terms.}
    \label{fig:failure}
    \vspace{-5pt}
\end{figure}

\subsection{LLM-Enhanced Generation and Fine-Tuning}
\label{sec:llm_enhanced_generation_and_fine_tuning}
Recent work has also integrated large language models (LLMs) to parse and refine prompts. LLM-grounded Diffusion~\cite{lian2024llmgrounded} and RPG~\cite{yang2024mastering} use LLMs to handle complex instructions, produce structured scene layouts, and plan multi-entity compositions. Though these frameworks enhance compositional consistency, they do not explicitly address subtle color nuances. Conversely, ColorPeel~\cite{butt2024colorpeel} proposes a fine-tuning strategy that learns a new prompt embedding for each specific color token, yielding accurate color rendering but at the cost of extra training. Our solution remains training-free yet leverages an LLM to disambiguate compound or ambiguous color terms, then directly blends their CIELab space representations within the diffusion process. By fusing LLM-based disambiguation strategies with precise color interpolation, we focus specifically on color accuracy, offering a lightweight alternative to training-heavy methods while expanding T2I models’ capability to handle intricate shades.

\section{Tint Benchmark}
\label{sec:tint_benchmark}

To estimate whether different T2I generation models can accurately interpret color descriptions, especially as users naturally employ expressive, context-dependent language, it is crucial to build a dataset that reflects real-world usage. Existing prompt datasets lack coverage of complex color expressions, focusing mainly on \textbf{basic color terms} like \textit{red} or \textit{blue}, and relying on rigid syntactic templates. This limits their utility in evaluating fine-grained color understanding. For instance, while CC-500 prompt dataset~\cite{feng2023trainingfree} explicitly defines attribute binding for color, its prompts follow fixed structures such as \textit{"a \{adj\} \{noun\} and a \{adj\} \{noun\}"} (\eg, \texttt{``a blue backpack and a red bench.''}), ailing to capture the richness of natural language. As analyzed in \cref{fig:failure}, enhancing prompts with more detailed and human-like descriptions often leads to generation failures in current T2I models. This highlights the need for a dataset that better reflects the expressive diversity of real-world color usage. To address these limitations, we introduce \textit{TintBench}, a benchmark that combines a curated taxonomy of compound color names, including blended, modified, object-based, signature, and abstract types, with prompts that resemble natural image captions. This design allows us to evaluate model performance in more realistic settings, where color expressions appear within diverse and contextually grounded descriptions. In the following sections, we describe how the compound color taxonomy is constructed and how the prompts are generated accordingly.

\begin{table}[]
    \centering
    \caption{Summary of the TintBench dataset. We report the number of prompts per category and provide representative examples after compound color substitution.}
    \resizebox{\linewidth}{!}{
    \begin{tabular}{>{\centering\arraybackslash}m{1.5cm} m{5.5cm}>{\centering\arraybackslash}m{1cm}}
    \toprule
    \textbf{Category} & \textbf{Example} & \textbf{Total}\\
    \midrule
    \textit{single-color} & \textit{"A woman with \textbf{ruby red} bags rides her bike over a bridge."} & 500 \\
    \midrule
    \textit{multi-color} & \textit{"A man with a \textbf{light blue} shirt and \textbf{Barbie pink} shorts is walking off of a soccer field."} & 500 \\
    \bottomrule
    \end{tabular}
    }
    \label{tab:tintbench}
    \vspace{-5pt}
\end{table}

\subsection{Compound Color Names}
\label{sec:compound_color_names}

Color perception varies significantly between individuals and is highly influenced by context. As a result, people use tens of thousands of phrases to describe colors. Among these, \textbf{compound color names} are especially important for expressing subtle shades and tones. A compound color typically consists of a basic color term-a foundational color category that can be modified with descriptive adjectives like \textit{light}, \textit{deep}, etc. To capture this diversity in color language, which is often overlooked in existing datasets, we curated an extensive collection of compound color names as a foundation for constructing TintBench. These compound color names provide the fine-grained semantics needed to rigorously test a model’s capacity for color understanding in realistic T2I scenarios.

Following prior work~\cite{berlin1991basic, moroney2024color}, we begin with eleven core basic colors: \textit{black}, \textit{blue}, \textit{brown}, \textit{gray}, \textit{green}, \textit{orange}, \textit{pink}, \textit{purple}, \textit{red}, \textit{white}, and \textit{yellow}. Building upon these, we searched the corresponding compound color names from web-based repositories such as Wikipedia and specialized color naming databases\footnote{For example, \url{https://colornames.org/}.}, and collected a diverse set of compound colors. Each compound color is linked to a standardized color representation (\textit{i.e.}, RGB code), enabling precise integration into prompt augmentation and evaluation.

The constructed compound colors can be categorized into the following five types:
\begin{itemize}
\item \textbf{Blended Color:} Created by combining two basic color terms to indicate a mixed hue, such as \textit{red purple} and \textit{yellow green}.
   \item \textbf{Modified Color:} Formed by modifying a basic color term with lightness-related adjectives, such as \textit{dark brown} and \textit{light blue}.
   \item \textbf{Object Color:} Constructed by prefixing a basic color term with the name of an object that represents the color, such as \textit{olive green} and \textit{salmon pink}.
   \item \textbf{Signature Color:} A color associated with a specific organization, institution, or geographical region, serving as an identifying hue, such as \textit{Duke blue} and \textit{Caribbean green}.
   \item \textbf{Abstract Color:} A color name where the prefixing adjective originates from abstract concepts, cultural references, or human-assigned labels, rather than physical attributes, such as \textit{Baker-Miller pink} and \textit{cyber yellow}.
\end{itemize}

By incorporating compound color names from all five categories, TintBench introduces challenging prompts that better reflect the complexity of real-world user inputs. The full list of compound color names and codes is available in the Appendix.

\subsection{TintBench Construction}
\label{sec:tintbench_construction}
Equipped with the collected compound colors, we aim to construct the prompt benchmarks in a natural way. Therefore, we selected the Flickr30k dataset~\cite{young2014image} as our starting point due to its relatively rich and varied descriptions. We first selected the prompts with color terms, leading to 30.21\% of the captions. Please note that these only contain basic color terms, and none sufficiently capture the diversity of compound color names. To introduce the compound colors, we first divided the captions into two groups: \textit{single-color} and \textit{multi-color}, based on the number of color terms present. We then applied \textit{k}-means clustering independently to each group, forming 20 clusters per group. From each cluster, we sampled five captions, resulting in 100 prompts for the \textit{single-color} group and 100 for the \textit{multi-color} group, for a total of 200 prompts.

Afterward, we augmented the selected prompts by replacing basic color terms with compound color names, randomly sampled from our five predefined categories. This process generated 500 \textit{single-color} and 500 \textit{multi-color} prompts, resulting in a total of 1,000 augmented examples. The augmentation enhances the granularity of color representation in the text prompts and enables more robust benchmarking of T2I models in interpreting descriptive, real-world color language. \cref{tab:tintbench} summarizes the composition of TintBench and provides representative examples.

\begin{figure}
\includegraphics[width=1.0\linewidth]{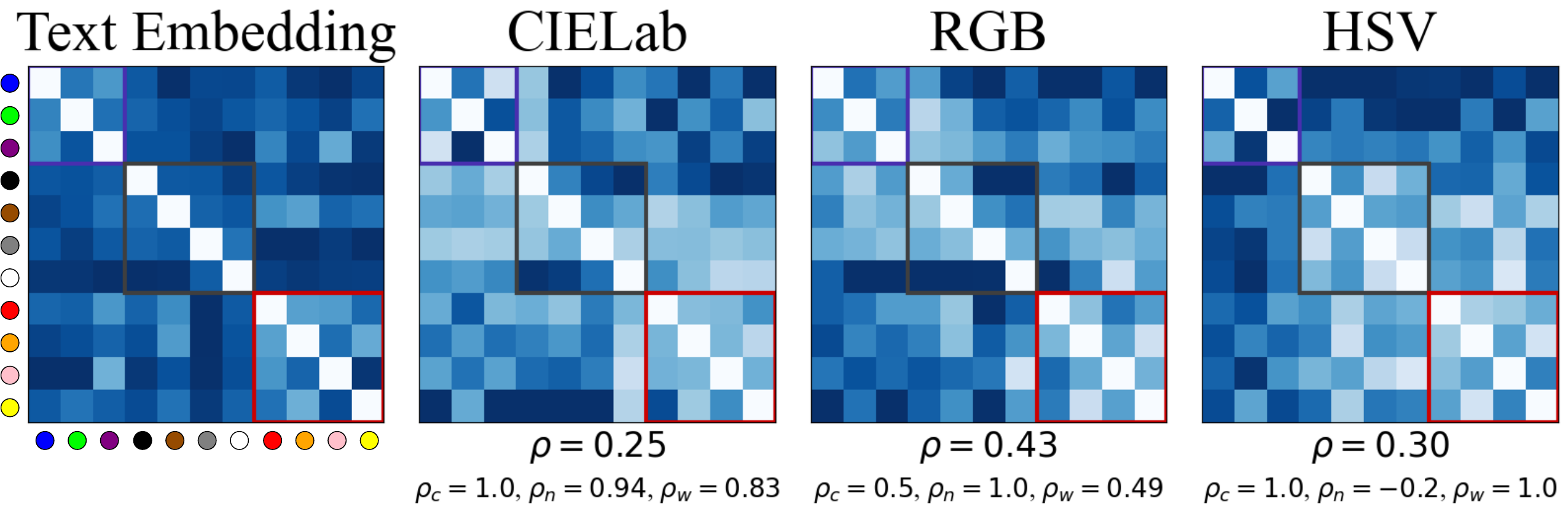}
    \caption{Comparison of pairwise distance matrices and Spearman correlations between text embeddings and color spaces. Overall correlation ($\rho$) is computed across eleven basic color terms, with group-wise coefficients ($\rho_w$, $\rho_n$, $\rho_c$) for warm (orange), neutral (gray), and cool (purple) colors. CIELab achieves the highest alignment overall and within all groups.}
    \Description{Comparison of pairwise distance matrices and Spearman correlations between text embeddings and color spaces.}
    \label{fig:all_color}
    \vspace{-10pt}
\end{figure}

\begin{figure*}
    \centering
    \includegraphics[width=1.0\linewidth]{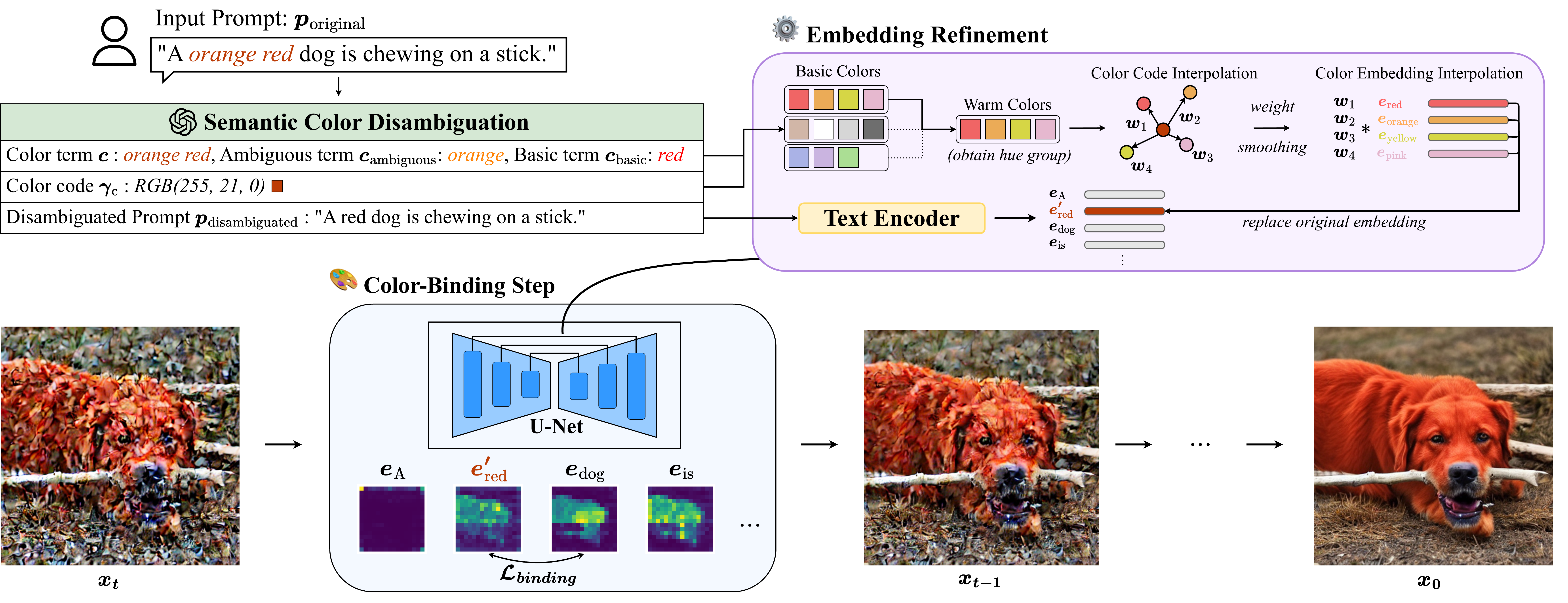}
    \caption{Our pipeline begins by using a large language model (GPT-4o) to resolve color ambiguity, producing a structured color analysis list. In the Embedding Refinement stage, the target color is interpolated with nearby basic colors within its Color Group to generate a precise embedding, which guides generation via cross-attention binding during denoising.}
    \Description{Our pipeline}
    \label{fig:pipeline}
\end{figure*}

\section{Method}
\label{sec:method}
Current methods often fail to resolve the ambiguity of diverse color terms and struggle to ensure precise color rendering, even when the intended semantics are correctly inferred. While fine-tuning with annotated color datasets is a potential solution, it is costly and requires expert input (e.g., via Photoshop). To address this, we propose a \textbf{training-free} framework for color blending in diffusion models via LLM-guided disambiguation, as illustrated in \cref{fig:pipeline}.

Given a text prompt with fine-grained or compound color expressions, our pipeline uses a large language model to rewrite ambiguous terms into clarified descriptions grounded in basic colors (\cref{sec:disambiguation}). Using CIELab codes of both the target and basic colors, we compute a color-space offset that captures their perceptual difference. This offset guides embedding refinement via interpolation in the text embedding space (\cref{sec:embedding_refinement}), analogous to semantic vector arithmetic (\eg, \textit{king} $-$ \textit{man} $+$ \textit{woman} $\approx$ \textit{queen}). The refined embeddings are then passed to the diffusion model, enabling more accurate and semantically faithful color generation.

\subsection{Semantic Color Disambiguation with LLMs}
\label{sec:disambiguation}
Text prompts often contain diverse compound color expressions, which can lead to the generation of unintended objects or visually inconsistent content. Recent T2I methods have incorporated POS-based analyses to improve attribute binding in diffusion models~\cite{rassin2023linguistic, feng2023trainingfree}. However, as shown in \cref{sec:tint_benchmark}, these approaches remain limited in handling semantically ambiguous or fine-grained color terms, often falling short of user expectations. Motivated by the recent progress of large language models (LLMs) in semantic understanding, we propose leveraging their strong reasoning capabilities to explicitly address the ambiguity in complex color descriptions.

As illustrated in the top-left of \cref{fig:pipeline}, our method employs an LLM (GPT-4o) to perform semantic analysis on text prompts containing intricate color expressions. Given an input prompt $p_\text{original}$, the LLM first analyzes each color term $c$. If an ambiguous term $c_\text{ambiguous}$ is identified—\textit{i.e.}, one likely to mislead the diffusion model—it is explicitly labeled. The LLM then selects the basic color term $c_\text{basic}$ that best represents the intended meaning and rewrites the prompt into a disambiguated version $p_\text{disambiguated}$ to resolve semantic confusion. In addition, the LLM provides a reference color code $\gamma_c$ (e.g., in RGB), informed by the scene context and intended interpretation of $c$. This process mitigates the risk of misinterpretation by aligning textual color semantics with perceptual expectations. The resulting color code directly informs the refinement of color-related embeddings, as detailed in \cref{sec:embedding_refinement}. Further details on prompt setup are provided in the Appendix.

\begin{table*}[t]
    \caption{User study results on TintBench. We report the average win rate for each evaluation metric. Bold values indicate a win rate exceeding 50\%.}
    \label{tab:user-study}
    \resizebox{\linewidth}{!}{
    \centering
    \def\arraystretch{1.1}
    \begin{tabular}{l*{4}{>{\centering\arraybackslash}p{1.6cm}}*{4}{>{\centering\arraybackslash}p{1.6cm}}}
    \toprule
    \multirow{2}[2]{*}{\textbf{Method}} & \multicolumn{4}{c}{\textbf{Single}} & \multicolumn{4}{c}{\textbf{Multiple}}\\
    \cmidrule(lr){2-5} \cmidrule(lr){6-9}
     & Prompt& Ambiguous& Color &Overall& Prompt& Ambiguous& Color& Overall\\
    \midrule
    SD~\cite{rombach2022high}& \textbf{95.84\%} & \textbf{60.44\%} & \textbf{79.17\%} & \textbf{93.75\%} & \textbf{86.11\%} & \textbf{66.67\%} & \textbf{86.11\%} & \textbf{94.44\%}\\
    AE~\cite{chefer2023attend}& \textbf{91.66\%} & \textbf{72.20\% }& \textbf{94.47\%} & \textbf{94.47\%} & \textbf{87.50\%} & \textbf{54.17\%} & \textbf{91.67\%} & \textbf{87.50\%}\\
    Conform~\cite{meral2024conform}& \textbf{83.35\%} & \textbf{62.50\%} & \textbf{75.00\%} & \textbf{79.20\%} & \textbf{63.89\%} & \textbf{75.00\%} &\textbf{80.55\%}& \textbf{66.66\%} \\
    DivideBind~\cite{li2023divide}& \textbf{75.00\%} & \textbf{66.67\%} & \textbf{69.44\%} & \textbf{80.56\%} &\textbf{74.90\%} & \textbf{68.06\%} &\textbf{68.54\%} & \textbf{83.34\%}\\
    Rich~\cite{ge2023expressive}& \textbf{77.78\%} & \textbf{69.44\%} & \textbf{86.11\%} & \textbf{83.33\%} &\textbf{75.00\%} & \textbf{75.00\%} & \textbf{68.75\%} & \textbf{83.34\%}\\
    InitNO~\cite{guo2024initno}& \textbf{77.09\%} & \textbf{79.17\%} & \textbf{68.75\%} & \textbf{79.16\%} &\textbf{86.07\%} & \textbf{74.97\%} & \textbf{86.90\%} & \textbf{91.53\%}\\
    SDG~\cite{agarwal2024training}&\textbf{87.70\%} & \textbf{79.17\%} & \textbf{85.42\%} & \textbf{89.58\%}&\textbf{79.17\%} & \textbf{79.17\%} & \textbf{68.75\%} & \textbf{83.33\%}\\
    SynGen~\cite{rassin2023linguistic}& \textbf{81.25\%}&\textbf{72.91\%} & \textbf{83.34\%}&\textbf{91.67\%} &\textbf{80.57\%}&\textbf{91.57\%}& \textbf{69.32\%}&\textbf{77.67\%}\\
    \midrule
    SDXL~\cite{podell2023sdxl}& \textbf{69.83\%} & \textbf{69.44\%} & \textbf{74.99\%} & \textbf{68.06\%}&\textbf{68.75\%}&\textbf{83.15\%}&\textbf{67.66\%}&\textbf{91.62\%}\\
    SDXL SynGen~\cite{rassin2023linguistic}&\textbf{81.48\%} & \textbf{74.99\%}&\textbf{84.26\%} & \textbf{87.96\%}& \textbf{79.16\%} & \textbf{79.16\%} & \textbf{69.47\%} & \textbf{87.50\%}\\
    SDXL Rich~\cite{ge2023expressive}&\textbf{88.10\%} & \textbf{84.50\%} & \textbf{65.00\%} & \textbf{89.28\%}  & \textbf{77.78\%} & \textbf{83.33\%} & \textbf{76.39\%} & \textbf{88.90\%} \\
    \bottomrule
    \end{tabular}
    }
    \label{tab:main}
    \vspace{5pt}
\end{table*}

\subsection{Retrieval-Based Embedding Refinement for Color Representation}
\label{sec:embedding_refinement}
Text embeddings encode semantic relationships between words in continuous vector spaces, enabling models to capture subtle linguistic similarities. Since the introduction of vector-based representations such as Word2Vec~\cite{mikolov2013efficient} and GloVe~\cite{pennington2014glove}, numerous studies have shown that semantically related concepts tend to occupy nearby regions in the embedding space. For example, embeddings of color terms like \textit{scarlet}, \textit{crimson}, and \textit{ruby}—all denoting shades of red—typically cluster together more closely than with unrelated terms such as \textit{blue} or \textit{green}. This compositional structure of text embeddings offers a promising foundation for modeling color semantics, particularly for interpolating between new color descriptions.

Motivated by these observations, we hypothesize that the spatial arrangement of color terms in learned text embedding spaces reflects their perceptual relationships in human color space. Specifically, we expect that terms representing similar hues (\eg, various shades of red) are embedded near each other, while those corresponding to perceptually distinct hues (e.g., \textit{red} vs. \textit{blue}) are more widely separated. To validate this hypothesis, we examine the spatial correlation between eleven basic color terms in the human color perception space and their corresponding representations in the text embedding space. 

Specifically, we first compute pairwise distances between color terms across several commonly used color spaces, including RGB, CIELab, HSV, YCbCr, YUV, and CIE1931. To account for the characteristics of non-linear color spaces, we adopt CIEDE2000 color difference formula ($\Delta E_{00}$)~\cite{sharma2005ciede2000} in the CIELab space. Unlike the simpler Euclidean formulation in CIE1931, $\Delta E_{00}$ incorporates corrections for lightness, chroma, and hue differences and includes a rotation term to account for perceptual interactions between chroma and hue in the blue region. The formula is defined as:

\vspace{3pt}
\begin{equation}
\resizebox{0.9\linewidth}{!}{%
$\Delta E_{00} = \bigg[ 
\left( \frac{\Delta L'}{k_L S_L} \right)^2 +
\left( \frac{\Delta C'}{k_C S_C} \right)^2 +
\left( \frac{\Delta H'}{k_H S_H} \right)^2
+ R_T \cdot \left( \frac{\Delta C'}{k_C S_C} \right)
\cdot \left( \frac{\Delta H'}{k_H S_H}\right)\bigg]^{1/2}$},
\label{eq:delta_e_00}
\end{equation}
\vspace{3pt}

where $\Delta L'$, $\Delta C'$, and $\Delta H'$ represent the differences in lightness, chroma, and hue, respectively; $S_L$, $S_C$, and $S_H$ are the weighting functions; $k_L$, $k_C$, and $k_H$ are typically set to $1$; and $R_T$ is a rotation term that accounts for the interaction between chroma and hue. For HSV, we isolate and project the hue component before computing distances. For all other linear color spaces, Euclidean distance is directly applied. We then calculate Spearman’s rank correlation coefficient ($\rho$) between each color space's distance matrix and the corresponding pairwise distances in the text embedding space. As shown in \cref{fig:all_color}, all color spaces exhibit only weak correlations with the text embedding space, suggesting that while linguistic and perceptual representations may share some structure, the alignment is imperfect and varies across encoding schemes.

Interestingly, we observe that the text embeddings of the color terms naturally form clusters based on hue categories. Based on this observation, we group the 11 basic color terms into three semantic categories:  \textbf{warm colors} (including \textit{red}, \textit{orange}, \textit{pink}, and \textit{yellow}), \textbf{cool colors} (\textit{blue}, \textit{green}, and \textit{purple}), and \textbf{neutral colors} (\textit{black}, \textit{white}, \textit{gray}, and \textit{brown}). We then compute Spearman correlations for each group individually. 

Results show that among all tested color spaces, CIELab consistently exhibits the highest positive correlation with the text embedding space across all three color groups, with an average correlation coefficient of 0.924. The correlation coefficients for the warm, neutral, and cool groups are denoted as $\rho_{w}$, $\rho_{n}$, and $\rho_{c}$, respectively, and are visualized in \cref{fig:all_color}. We attribute this outcome to findings in color science, which suggest that human vision is particularly sensitive to variations in lightness. As a perceptually uniform space, CIELab more closely reflects how humans perceive color differences in images. This result also aligns with findings from ColorPeel~\cite{butt2024colorpeel}, which showed that projecting diffusion latents into CIELab enables more precise control over generated colors. Consequently, we adopt CIELab as the default color space for all subsequent operations.

To further demonstrate that the text embeddings can be manipulated to achieve more accurate representations of target colors, we perform color blending within embedding space based on the relationships among basic color terms in perceptual color space. Specifically, we apply proportional blending to the text embeddings of color terms, following the behavior of color interpolation in the CIELab space. For example, the Lab code (66, 43, 68) corresponds to a darker shade of orange, while (92, -21, 94) represents a highly saturated yellow. Averaging these two values yields a color code that lies between them, producing a yellow-orange hue that visually resembles a perceptual blend of both original colors. 

\begin{figure}
    \centering
    \includegraphics[width=1.0\linewidth]{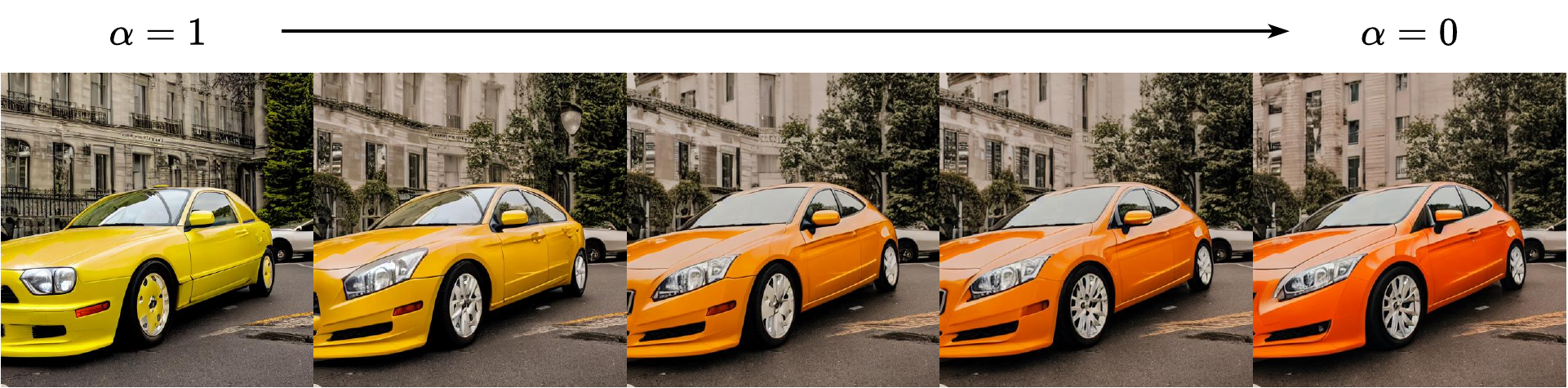}
    \caption{Interpolated text embeddings between \texttt{``orange car.''} and \texttt{``yellow car.''} As the blending ratio changes, the resulting hues transition smoothly from orange to yellow, depending on the dominant color.}
    \Description{Interpolated text embeddings}
    \label{fig:cars}
    \vspace{-15pt}
\end{figure}

Following this principle, we interpolate between the text embeddings of orange and yellow in varying proportions, controlled by a blending factor $\alpha \in [0,1]$. Specifically, we compute the interpolated embedding using the formula:
$\alpha \cdot e_{\text{yellow}} + (1 - \alpha) \cdot e_{\text{orange}}$,
where $e_{\text{yellow}}$ and $e_{\text{orange}}$ denote the text embeddings of the tokens \textit{orange} and \textit{yellow}, respectively. The resulting vector is then used to replace the color token embedding in the prompt.
As shown in \cref{fig:cars}, when the blending ratio $\alpha$ shifts, the car's color also shifts toward either orange or yellow, depending on the dominant component.

\begin{figure*}[t]
    \vspace{10pt}
    \centering
\includegraphics[width=1.0\linewidth]{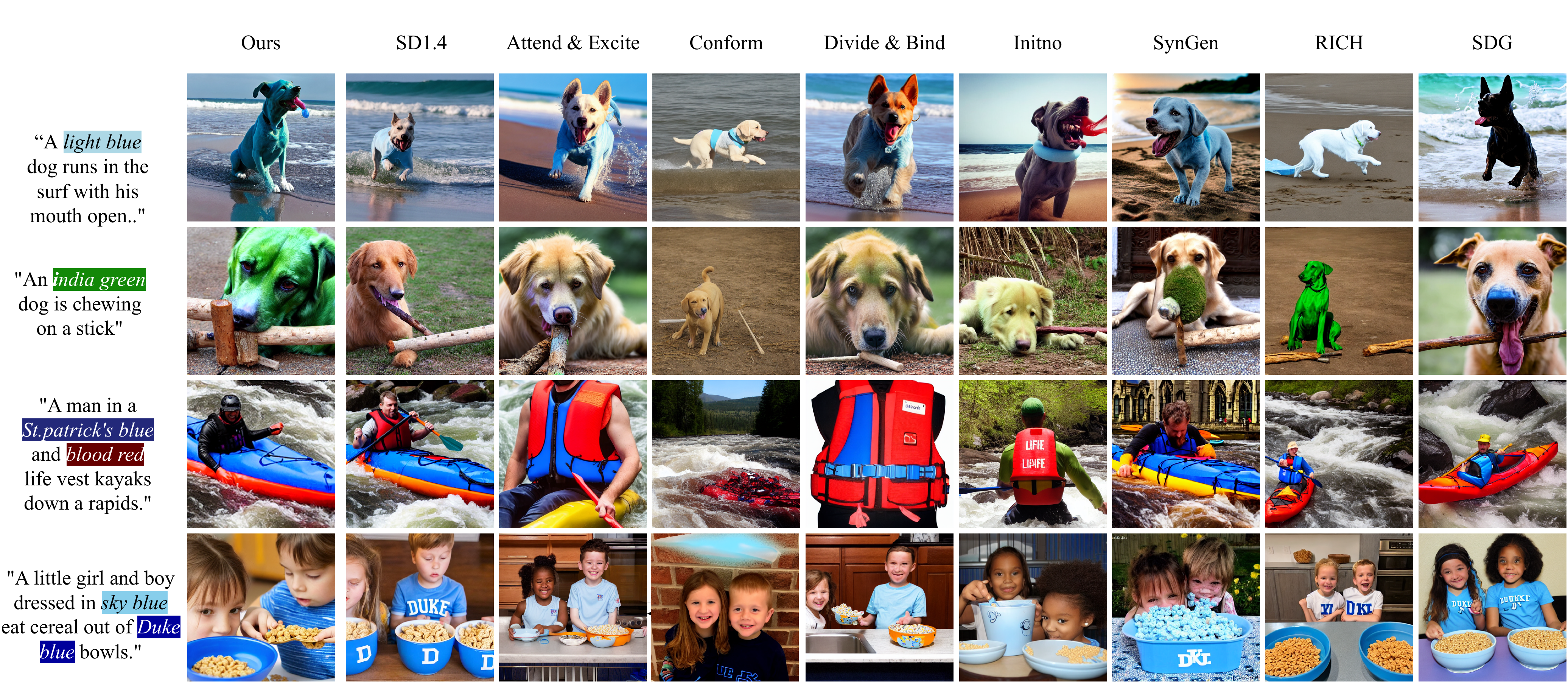}
    \caption{Qualitative comparison between our method and previous work using different prompts in SD v1.4. The first two rows show results for single-color prompts, while the last two rows demonstrate the multi-color prompt setting.}
    \Description{Qualitative comparison}
    \label{fig:exp_qualitative}
    \vspace{10pt}
\end{figure*}
 
To enable perceptually grounded color blending, we first classify the predicted color into a predefined hue group based on the color code generated by the LLM during the \textbf{Semantic Color Disambiguation} stage. As demonstrated in our earlier analysis, hue groups exhibit a stronger correlation between perceptual color distances and text embedding similarities. Restricting subsequent operations to colors within the same hue group ensures higher semantic relevance in the blending process. Within the identified hue group, we then retrieve the top-$k$ nearest basic color terms by computing perceptual distances in the CIELab space using the $\Delta E_{00}$ metric. This metric closely aligns with human visual perception and provides a more reliable measure than Euclidean distance in RGB or other non-uniform color spaces. Prior work~\cite{moroney2024color} has shown that diffusion models respond more consistently to basic color terms. Motivated by this, we treat the embeddings of the selected basic color terms as directional anchors, using them to guide the offset computation for the target color term in the text embedding space. This strategy ensures that the blended embedding is grounded in semantically and perceptually salient color representations. To compute the blending weights between the selected basic color embeddings, we move beyond simple linear interpolation and instead use a Gaussian-based softmax formulation. This accounts for the non-linear structure of the CIELab space and yields smooth, perceptually meaningful blending. The target color embedding $e_\text{target}$ and weight $\alpha_i$ for each basic color $i$ is defined as:

\begin{equation}
e_{target} = \sum_{i}^{k} \alpha_i e_i, \ \text{with} \  \alpha_i = \text{softmax}\left(-\frac{d_i^2}{2\sigma^2}\right),
\label{eq:gaussian_blending}
\end{equation}

where $k$ is the number of color terms, $e_i$ is the embedding for the basic color $i$, $d_i$ is the perceptual distance ($\Delta E_{00}$) between the target color and the $i$-th basic color, and $\sigma$ is a temperature-like scaling factor that controls the sharpness of the weight distribution. Through this retrieval mechanism, we can generate text embeddings that accurately capture the semantics of the blended color representation. This ensures that the resulting embedding effectively conveys the intended color meaning.

To improve the binding between color terms and their corresponding visual entities, we incorporate a guidance loss during the denoising step. This component is inspired by the positive loss formulation in Cross-Attention-based Guidance used in SynGen~\cite{rassin2023linguistic}. Specifically, the Color-Binding loss $\mathcal{L_\text{binding}}$ is applied after embedding refinement and serves as a soft constraint that guides the attention maps toward more semantically consistent color-object alignment:

\begin{equation}
\mathcal{L}_{\text{binding}}^i = \frac{1}{2} \mathcal{D}_{\mathrm{KL}}(A_\text{color}^i \parallel A_\text{entity}^i) + \frac{1}{2} \mathcal{D}_{\mathrm{KL}}(A_\text{entity}^i \parallel A_\text{color}^i),
\label{eq:symmetric_kl}
\end{equation}

where $A_{\text{color}}^i$ and $A_{\text{entity}}^i$ denote the cross-attention maps corresponding to the $i^{\text{th}}$color term and the $i^{\text{th}}$ entity term, respectively. Each attention map is normalized such that its elements sum to 1. The symmetric Kullback-Leibler divergence $\mathcal{L}_{\text{binding}}$ encourages alignment between these two distributions, guiding the model to associate the correct color with the correct visual region. After computing the binding loss $\mathcal{L_\text{binding}}$, we update the output latents $x_t$ during the denoising step $t$ by the \textbf{Color-Binding Step}:

\begin{equation}
x_t \leftarrow x_t - \alpha \cdot \nabla_{x} \sum_{c \in C} \mathcal{L}_\text{binding}^c,
\label{constraint}
\end{equation}

where $\alpha$ is a scalar that controls the binding scale, and $C$ represents the set of basic color terms identified in the input prompt.
\section{Experiments}
\label{sec:experiments}
\subsection{Experimental setup}

\noindent \textbf{Dataset.} We evaluate our method using \textit{TintBench}, a benchmark we constructed as described in Section~\ref{sec:tint_benchmark}, which includes 500 \textit{single-color} and 500 \textit{multi-color} prompts. These prompts cover five types of compound color expressions, as previously outlined.

\noindent \textbf{Implementation Details.} We run our model on both Stable Diffusion (SD) 1.4 and SDXL. Image generation is performed using the DDIM scheduler with 50 denoising steps per image. Experiments with SD 1.4 are conducted on NVIDIA RTX 3090 GPUs, while those with SDXL are conducted on NVIDIA V100 GPUs.

\noindent \textbf{Comparison Baselines.} To evaluate the effectiveness of our approach, we compare it with several representative baseline methods that primarily rely on attribute binding and cross-attention control. Specifically, we include Attend-and-Excite~\cite{chefer2023attend}, Conform~\cite{meral2024conform}, DivideBind~\cite{li2023divide}, Rich~\cite{ge2023expressive}, Initon~\cite{guo2024initno}, SDG~\cite{agarwal2024training}, and SynGen~\cite{rassin2023linguistic}.

\subsection{Quantitative Result}
\label{sec:quantitative}

Following prior work, we assess the quality and fidelity of generated images through a comprehensive user study, where participants compare outputs from our method against several strong baseline approaches. The results are summarized in \cref{tab:user-study}. The evaluation focuses on three key aspects: (1) \textbf{Prompt Alignment}, measuring how well the generated image matches the full textual prompt; (2) \textbf{Color Fidelity}, assessing the accuracy and realism of the rendered colors; and (3) \textbf{Ambiguity Resolution}, evaluating the method's ability to handle context-sensitive or vague color descriptions that often lead to misinterpretation or visual inconsistencies. As shown in the table, our method consistently achieves the highest average win rates across nearly all metrics, with bold numbers indicating win rates exceeding 50\%. This demonstrates that users significantly prefer our results over those of other methods. Notably, our performance is particularly strong in the “Ambiguous” category, highlighting the effectiveness of our semantic disambiguation strategy. These results underscore our method’s advantage in understanding fine-grained and nuanced color expressions, which are challenging for most existing baselines. In addition, we also demonstrate strong performance when integrating our pipeline with SDXL, further validating the robustness and flexibility of our approach in both standard and high-capacity diffusion settings. Additional analyses on experimental results are provided in the Appendix.

\subsection{Qualitative Result}
\begin{figure}
    \centering
\includegraphics[width=1.0\linewidth]{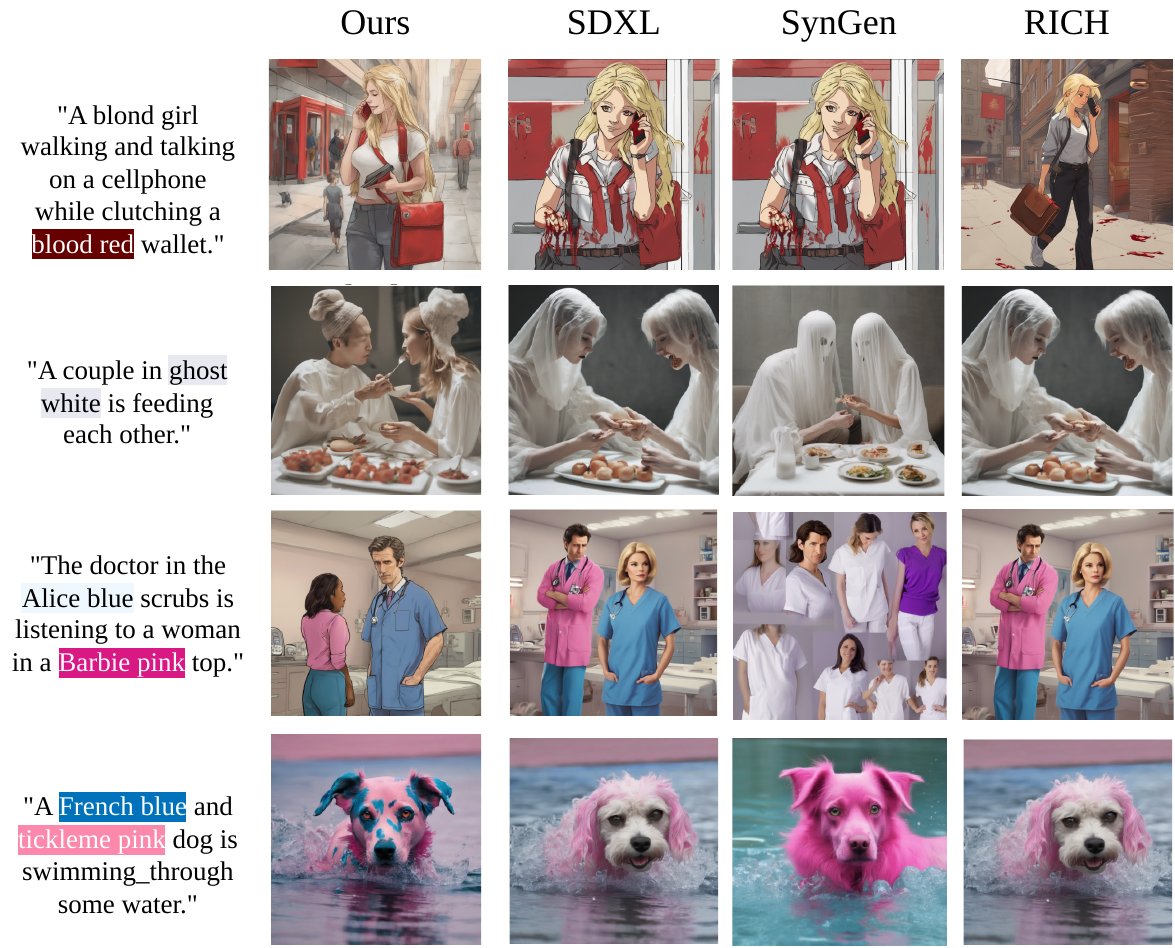}
    \caption{Qualitative comparison between our method and previous work using different prompts in SDXL. The first two rows show results for single-color prompts, while the last two rows demonstrate the multi-color prompt setting.}
    \Description{Qualitative comparison}
    \vspace{-15pt}
\label{fig:appendix_qualitative_sdxl}
\end{figure}

We present a qualitative comparison between our method and other approaches in~\cref{fig:exp_qualitative} and~\cref{fig:appendix_qualitative_sdxl}. In~\cref{fig:exp_qualitative}, we show that the original Stable Diffusion 1.4 (SD1.4) model is easily misled by ambiguous color terms in the prompt. For example, in row 4, the word "DUKE" appears on the shirt of a young boy—an artifact caused by the model interpreting the compound color name "\textit{Duke blue}" as referring to the university rather than the intended color. Although previous works on attribute binding help mitigate this issue to some extent, they still struggle to generate accurate compound colors as defined in our Tint Benchmark. In contrast, our method addresses both the ambiguity in prompts and successfully generates the correct color output according to the specified compound color. Similarly,~\cref{fig:appendix_qualitative_sdxl} demonstrates that this issue also occurs in Stable Diffusion XL (SDXL), and again, our method effectively resolves it. These results highlight the importance of semantic color disambiguation when interpreting compound color names. By integrating perceptual grounding and LLM-driven understanding, our approach ensures that visually grounded color meanings are preserved during generations.

\section{Conclusion and Future Work} \label{sec:conclusion}

In this work, we addressed the long-overlooked challenge of fine-grained color understanding in text-to-image generation. We introduced TintBench, a comprehensive benchmark constructed by augmenting real-world prompts with diverse and nuanced compound color expressions. Our dataset enables a more rigorous evaluation of how well generative models capture human-like interpretations of color semantics. To bridge the gap between textual color descriptions and perceptual color representations, we proposed a training-free pipeline that leverages LLMs for semantic color disambiguation. Additionally, we introduced a perceptually weighted interpolation mechanism grounded in the $\Delta E_{00}$ metric, enabling smooth and semantically meaningful blending of basic color embeddings. Our user studies and visualizations demonstrate that the proposed approach offers superior color fidelity and effectively resolves color ambiguities in prompt interpretation. We hope that TintBench and our accompanying framework will facilitate future research in controllable generation and semantic grounding, particularly in applications that demand high-fidelity coloring. Looking ahead, we plan to scale our dataset to include a wider range of input modalities, such as raw color codes, image regions, and free-form text descriptions. We also aim to develop a foundation model for color grounding that can flexibly interpret and generate accurate color representations across modalities. This would pave the way toward more generalizable and interactive multimodal generation systems with fine-grained color control.

\clearpage

\balance
\begin{acks}
This work is partially supported by the National Science and Technology Council, Taiwan, under Grant: NSTC-112-2221-E-A49-059-MY3 and NSTC-112-2221-E-A49-094-MY3.
\end{acks}

\bibliographystyle{ACM-Reference-Format}
\bibliography{main}


\end{document}